\documentclass[sigconf]{acmart}

\AtBeginDocument{%
  \providecommand\BibTeX{{%
    \normalfont B\kern-0.5em{\scshape i\kern-0.25em b}\kern-0.8em\TeX}}}

\setcopyright{acmcopyright}
\copyrightyear{2022}
\acmYear{2022}
\acmDOI{XXXXXXX.XXXXXXX}

\acmPrice{0.00}
\acmISBN{978-1-4503-XXXX-X/18/06}

\begin{document}

\title{Testing the Robustness of a BiLSTM-based\\ Structural Story Classifier}


\author{Aftab Hussain}
\affiliation{%
  \institution{University of Houston}
  \city{Houston, Texas}
  \country{USA}}
\email{ahussain27@uh.edu}

\author{Sai Durga Prasad Nanduri}
\affiliation{%
  \institution{University of Houston}
  \city{Houston, Texas}
  \country{USA}}
\email{snanduri4@uh.edu}

\author{Sneha Seenuvasavarathan}
\affiliation{%
  \institution{University of Houston}
  \city{Houston, Texas}
  \country{USA}}
\email{sseenuvasavarathan@uh.edu}

\renewcommand{\shortauthors}{Hussain et al.}

\begin{abstract}
The growing prevalence of counterfeit stories on the internet has fostered significant interest towards fast and scalable detection of fake news in the machine learning community. While several machine learning techniques for this purpose have emerged, we observe that there is a need to evaluate the impact of noise on these techniques’ performance, where noise constitutes news articles being mistakenly labeled as fake (or real). This work takes a step in that direction, where we examine the impact of noise on a state-of-the-art, structural model based on BiLSTM (Bidirectional Long-Short Term
Model) for fake news detection, Hierarchical Discourse-level Structure for Fake News Detection by Karimi and Tang~\cite{karimi}. 
\end{abstract}

\keywords{Model robustness testing, structural learning}

\maketitle

\section{Introduction}
\label{sec-intro}

With the growth of social media, such as Twitter and Facebook, and various online news media outlets, the widespread reach of fake news has become a very detrimental problem for society, triggering the growth of numerous online platforms for fact-checking. Towards checking counterfeit content, a great deal of interest as also grown in the Machine Learning (ML) field from where various software systems for detecting fake content have emerged over the past few years. Given the vast datasets that are used to train such ML-based systems, we underscore that there is a need to evaluate the robustness of these systems in the presence of noise. 

Driven by question, \textit{``Are ML models in the domain of counterfeit news detection smart enough to bypass perturbations in the training data and make quality predictions based on what they learn from the structure of the data?"}, in this work, we evaluate the impact of noise on a recent ML technique used in fake news detection: Karimi and Tang’s Hierarchical Discourse-level Structure for Fake News Detection (HDSF) framework~\cite{karimi}. In particular, we explore how the prediction accuracy of HDFS is impacted when HDFS is trained with noise-induced datasets? Here, we focus on the problem where noise entails news article data points in the training set being mis-labelled as fake or real.   

The framework we choose for evaluation, HDFS, is based on Bi-directional Long-Short-Term Models, that learns the context of news documents by capturing structural information of the constituent sentences of the documents. It has been found to outperform other standard models (e.g. N-Grams) in the field of fake news detection. In addition, it has been shown that robust BiLSTM models, the building block of HDFS, can be built in the domain of disfluency detection in speech transcripts~\cite{bach19_interspeech}. Our work takes a stepping stone in the direction of determining how accurate are the predictions of the ML models that have been trained with noisy data in the domain of fake content detection. 

\noindent \textbf{Contributions.} Overall we make the following contributions in this paper:
\begin{itemize}
    \item We present an empirical evaluation of the impact of noise on the Hierarchical Discourse-level Structure for Fake News Detection (HDSF) machine learning model proposed by Karimi and Tang~\cite{karimi}.
    \item We provide insights into how evaluations of fake news detection machine learning systems can be done to understand their learning behaviour.
\end{itemize}

\noindent \textbf{Paper Organization.} In Section~\ref{sec-back}, we describe the foundations of HDSF. Then in Section~\ref{sec-method}, we elaborate on our methodology for evaluating the impact of noise on HDSF, where we discuss how we introduced noise in the train sets, and show the workflow of how we evaluated HDSF. In the subsequent sections, we describe the settings of our experiments (Section~\ref{sec-settings}), present our experimental findings (Section~\ref{sec-results}), and provide a discussion on our evaluation (Section~\ref{sec-disc}). In Section~\ref{sec-related}, we elaborate upon some related works. We conclude our paper in Section~\ref{sec-conclusion}.
\section{Background on HDSF}
\label{sec-back}

The core motivation of the Hierarchical Discourse-level Structure for Fake news detection framework (HDSF)~\cite{karimi} is to capture the structure of news articles and exploit this structural information for determining fake news. In addition, HDSF provides a path towards better understanding the language of fake news. Two observations, as per different studies, primarily drove their work towards extracting hierarchical structures in analyzing news: (1) Typically, fake news is generated by combining separate, disjoint news parts, and (2) using hierarchical structures for document representations produce better results in prediction tasks (where the entire text is treated as the main predictor variable). Their work was found to outperform other baseline models such as N-grams, RST~\cite{truth-deception}, BiGRNN-CNN (Bi-directional General Neural Network-Convolutional Neural Network), and LSTM (Long Short Term Memory) in accuracy.

HDSF is based on the BiLSTM (Bidirectional Long-Short Term Model)~\cite{phoneme,birnn}. BiLSTM is a sequence processing model consisting of two long-short-term models, the first one learns the input sequence and the second one captures the reverse of the same sequence. The inputs are taken in opposite directions by each model (the first one takes forward, and the other takes the input in backward direction). This approach in BiLSTMs allows them to store more information about the input in the network, and thereby pick up more contextual information about the input~\cite{wiki-bilstm}. 

\begin{figure*}
  \includegraphics[width=\textwidth]{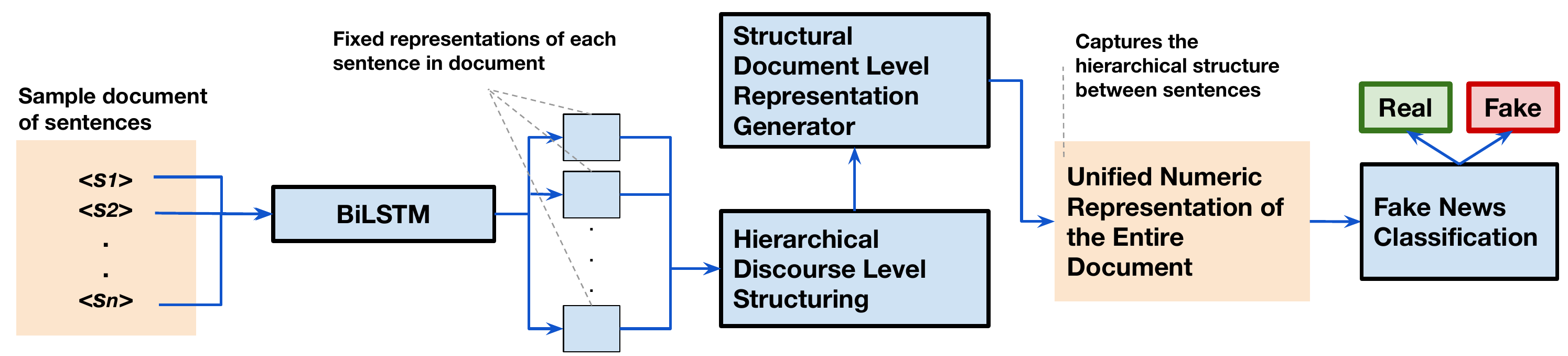}
  \caption{The HDSF framework~\cite{karimi}.}
  \label{fig:hdfs}
\end{figure*}

In Figure~\ref{fig:hdfs}, we give a simplified re-representation of the framework of HDSF when used on a single sample document~\cite{karimi}. The framework first obtains a fixed representation (a fixed-size word embedding) of each sentence in a document using a Bi-directional Long-Short Term Memory network. Next, these sequence of sentence embeddings are fed into the Hierarchical Discourse-level Structure Learning component, which constructs a dependency tree of the sentences in the document by using a dependency parsing approach~\cite{liu-lapata-2018-learning}. The tree encapsulates semantic dependency relationships between sentences. The entire tree of the document is then output in a unified numeric format representation. This representation of the document is then passed on to the Fake News Classifier which is a binary classifier that utilizes probabilities of the document being fake or real to compute a cross-entropy loss value. For brevity, details of the computation of the loss function have been omitted (they are discussed in~\cite{karimi}).

\section{Methodology}
\label{sec-method}

\begin{figure}
  \includegraphics[width=85mm]{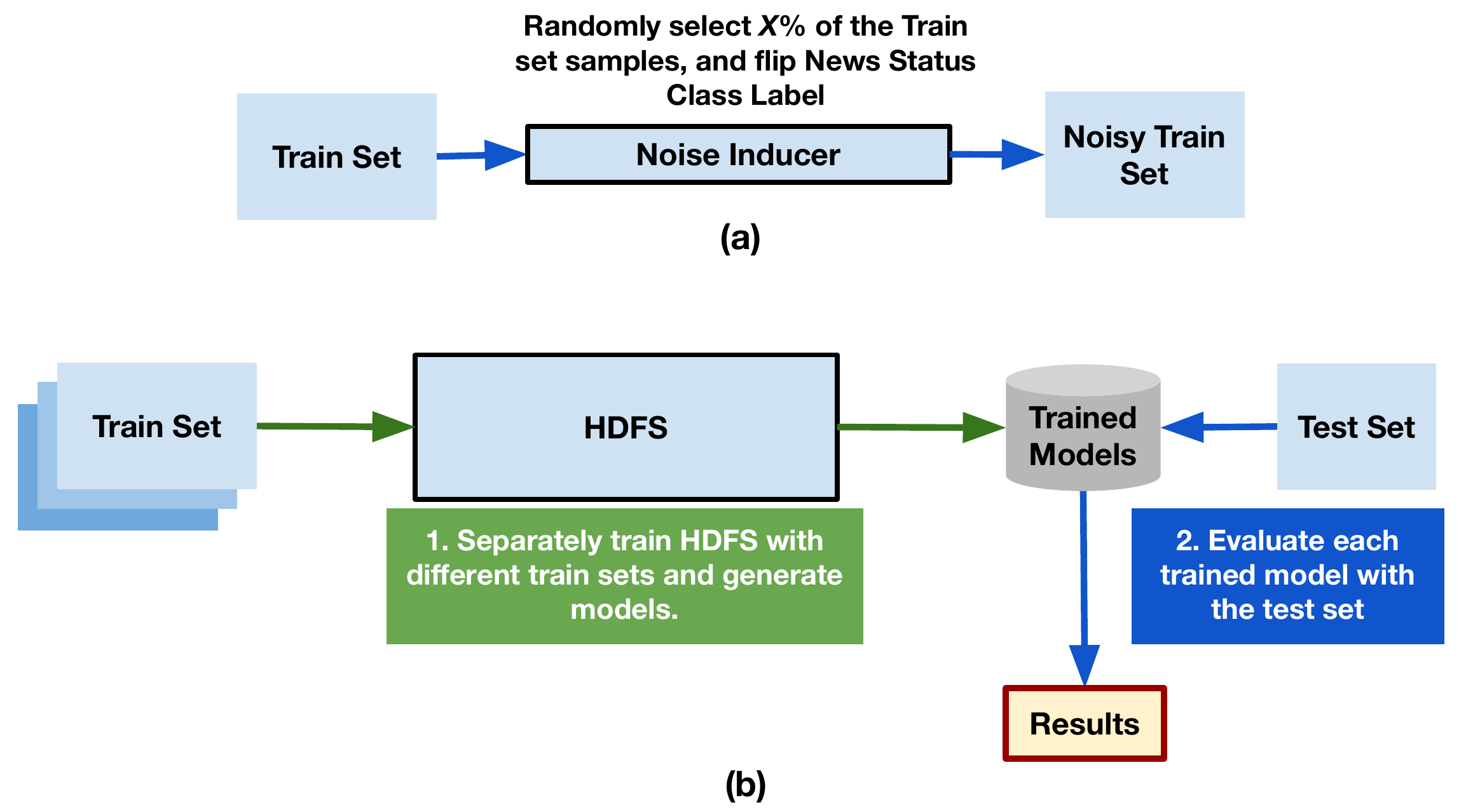}
  \caption{Overview of our approach for evaluating the impact of training HDFS with noisy datasets. (a) Approach for the generation of noise induced datasets. (b) Training HDFS with datasets, and evaluating it with the test set.}
  \label{fig:noise-inducer}
\end{figure}

In this section, we present our approach for evaluating the impact of noise on HDSF. We first describe the noise generation process in Subsection~\ref{subsec-noisegen}. Then, we elaborate on the training process in Subsection~\ref{subsec-train}. Figure~\ref{fig:noise-inducer} provides an overview of the entire approach.

\subsection{Generation of Noise-Induced Datasets}
\label{subsec-noisegen}

We first produce the noisy datasets using our Noise Inducer module (Figure~\ref{fig:noise-inducer} (a)), which is written in Python. The Noise Inducer first shuffles the train set and randomly picks a sample of the training set with 50\% probability, and flips its news status or authenticity label: changing ``Fake" to ``Real", and vice versa. The Noise Inducer also keeps track of which samples have been already changed, to avoid re-altering an already altered sample. The Noise Inducer repeats this process on the training set until $X$\% of the samples have been modified, where $X$ is a user-defined parameter. In this manner, we generated four different levels of noise-induced datasets from the original dataset: training sets at 25\%, 50\%, 75\%, and 100\% noise.

\subsection{Training HDFS}
\label{subsec-train}

Once the train sets have been generated, we deploy HDSF available in HDSF's authors' Github repository~\cite{hdfs-repo}. We trained HDSF with each of the five training sets (one original and four noisy dataset) over 20 steps, and generated models, on which we performed our tests using a separate test set (Figure~\ref{fig:noise-inducer} (b)). Thereafter we collected loss and accuracy scores for each model obtained for the same test set.

\section{Experimental Settings}
\label{sec-settings}

In this section we describe the settings we used to conduct our experiments. In Subsection~\ref{subsec-ds}, we describe the properties of the datasets we used. In Subsection~\ref{subsec-setup}, we detail the configuration of our system and the HDFS model that we deployed.

\subsection{Datasets}
\label{subsec-ds}

We use the same dataset to train the HDSF model as that used in~\cite{karimi}. In total, the dataset consists of 6,452 examples. For investigating the impact of noise on HDSF, we generated four different noisy datasets from this dataset using the approach mentioned in Section 3 at noise levels of 25\%, 50\%, 75\%, and 100\%, respectively. All datasets have the same number of samples. After training, we tested each obtained model with a test set of 134 samples.

\subsection{Setup}
\label{subsec-setup}

We ran the experiments on a system with 12 Intel(R) Xeon(R) Bronze 3204 1.90GHz processors with 62 GB RAM, running a Ubuntu 18.04.5 LTS operating system. The HDFS model was trained for 200 steps with each training set. We used the default HDFS model settings, as used in the original repository~\cite{hdfs-repo}: batch size of 40 samples, BiLSTM hidden unit dimension of 100, and drop-out rate of 0. The total execution time for running all the training sessions was around 20 and a $\frac{1}{2}$ hours, taking roughly four hours per training session.

\section{Results}
\label{sec-results}

In this section, we present the results of our experiments, where we evaluate the impact of noise on HDFS. In particular, we address the following research questions:

\begin{itemize}
    \item[RQ.1] What is the impact of dataset noise on the training loss of HDFS? (Subsection~\ref{subsec-rq1})
    \item[RQ.1] What is the impact of dataset noise on the prediction accuracy of HDFS? (Subsection~\ref{subsec-rq2})
\end{itemize}

\subsection{Impact of Dataset Noise on Training Loss}
\label{subsec-rq1}

\begin{figure*}
  \includegraphics[width=150mm]{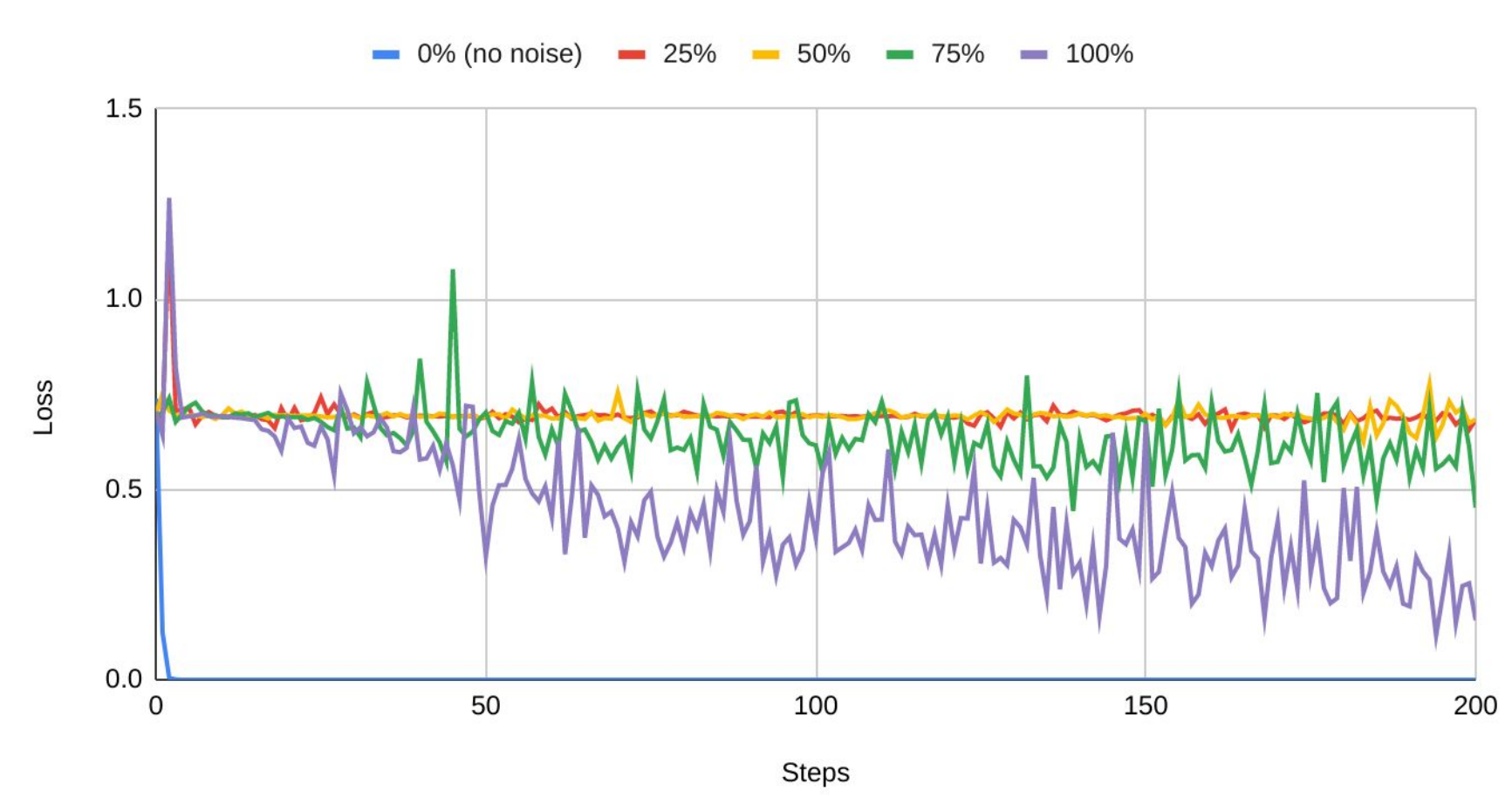}
  \caption{HDFS training loss when trained with different levels of noise-induced datasets.}
  \label{fig:hdfs-trainloss}
\end{figure*}

Figure~\ref{fig:hdfs-trainloss} shows HDFS’s loss scores obtained while training it with the five different datasets (original/no-noise data, 25\%-, 50\%-, 75\%-, 100\%-noise induced data). We see that over the 200 steps of training, the training loss is the lowest with the original dataset. For the other four datasets we see three interesting trends: (1) The training loss is around 0.7 when HDFS is trained with the 25\% and the 50\% noise-induced datasets, throughout the training process, (2) the training loss is more erratic with the 75\% and 100\% noise-induced datasets, with the 100\% noise-induced dataset incurring less loss than all the other noisy datasets, and (3) as the number of steps increases, training with the 100\%-noise induced dataset shows improvement in the loss score, unlike when using the other noise-induced datasets, where the training losses tend to hold a constant horizontal trend.

\vspace{5pt}

\noindent \textbf{Observation.} \textit{While training with the original (noise-less) dataset exhibited the lowest training loss, adding more noise to the dataset does not increase the loss in training in accordance with how much noise we add to the dataset.}

\subsection{Impact of Dataset Noise on Prediction Accuracy}
\label{subsec-rq2}

\begin{figure*}
  \includegraphics[width=175mm]{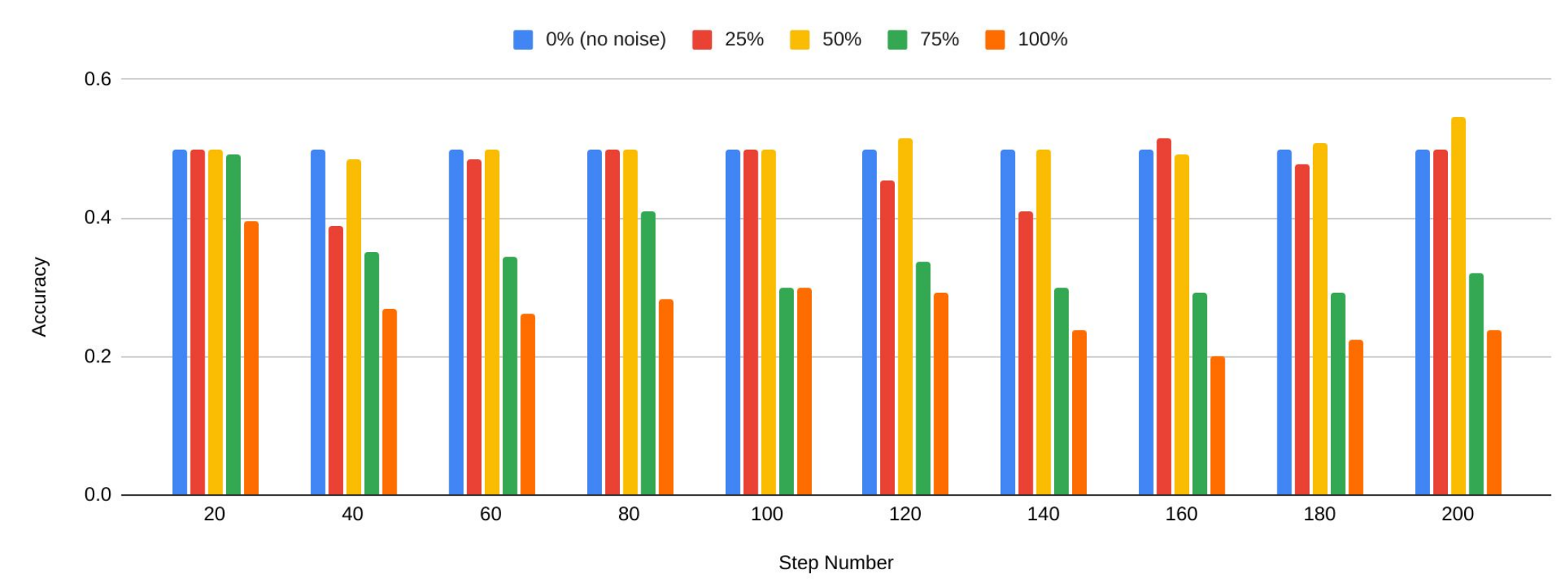}
  \caption{Test Accuracy of the HDFS Model at 10 different stages of training, for different levels of noise-induced datasets.}
  \label{fig:hdfs-test}
\end{figure*}

 In Figure~\ref{fig:hdfs-test}, we compare the final accuracies of the model obtained at each of 10 different stages of the training process, for each of five different versions of the dataset. We observe that the prediction accuracy of the HDFS model on test data remains the same at different stages of the training process, when the model is being trained with the untampered dataset. With the noise-induced datasets, we see several trends: (1) At several stages, the performance with the noise-induced datasets are as good as that with the original dataset, (2) at certain stages, HDFS trained with some of the noisy datasets outperformed the model when trained with the original dataset (for e.g. the model trained by the 50\% noise induced dataset at steps 120, 180, and 200, and the model trained by the 25\% noise induced dataset at step 160), and lastly (3) the fluctuations in performance from stage-to-stage are more pronounced with the noise-induced datasets than that with the original dataset. 
 
 \vspace{5pt}
 
 \noindent \textbf{Observation.} \textit{Adding high-levels of noise (100\% and 75\%) significantly reduced HDFS's prediction accuracy on unseen (test) data, however, HDFS made better predictions for some moderate noise levels (25\% and 50\%) at certain stages than it did when trained with the original dataset.}

\section{Discussion}
\label{sec-disc}

From the experiments we observe that noise can have an unexpected effect on the accuracy of the model – as such increasing noise may not always decrease the accuracy of the model. We believe this can exhibit the generalizability~\cite{arpit} potential of HDFS, whereby it is relying on the structure of the code, rather than blindly following the labels in the training set – which is a positive sign as it conversely shows the model is relying less on memorizing~\cite{arpit} the data. However, there is a limit as to how much noise the model can take. For example, we observed that the accuracy of the model drops significantly when trained at a 100\% noise level. Thus, further investigations are necessary to investigate the memorization and generalization behaviour of the model. Lastly, in our experiments the size of the test set was relatively very small (134 samples). A larger variety of test samples may be tried to obtain more generality in the results. It also remains to be seen whether the trends revealed in these experiments can extend to other datasets.

\section{Related Works}
\label{sec-related}

There have been several studies that evaluated the impact of noise on machine learning systems in various domains such as image classification (e.g.~\cite{arpit,still-rethink-gen}), code intelligence (e.g.~\cite{rabin2021memorization}), and others. To the best of our knowledge, there have not been similar works for models in the domain of counterfeit news detection. Horne et al.~\cite{horne} evaluate the robustness of a fake news classifier in the presence of specific adversarial attacks in the input space, however, they do not analyze the behaviour of the model with randomized noise, as was done in this work. Below we discuss some contributions by the ML community in the domain of fake content detection.

Hakak et al. ~\cite{hakak} aim to improve training and testing accuracy by proposing an ensemble-based model. The model, built on Random Forest, Decision Tree, and Extra Tree Classifier, focuses on extracting important features from the dataset. In total, they extracted 26 features including number of words, number of characters, number of sentences, average sentence length and others. They implement their approach on the LIAR dataset~\cite{wang2017liar}. Aslam et al.~\cite{fdetect} also proposed an ensemble-based machine learning model that combines two deep learning models based on the properties of attributes: Bi-LSTM-GRU-dense deep learning model for a textual attribute (statement) and a dense deep learning model for other attributes. They also apply their approach on the LIAR dataset. 

Liu and Wu tackle the problem of fake news detection from the angle of early detection ~\cite{fned}. They propose the notion of crowd response for a news article, which consists of a sequence of user responses. They assume, the larger the sequence of responses, the more the news has propagated. Their ML technique they propose is a deep neural network that uses a Convolutional Neural Network (CNN) based classifier. They implement their approach on Twitter and Weibo datasets~\cite{ma-etal-2018-rumor, microblogs}. An earlier work by Zhou et al.~\cite{fake-news-theory} also focuses on the problem of early fake news detection, but applies a theory based approach. They investigate news content at multiple levels including, lexicon-level, syntax-level, semantic-level, and discourse-level, where well-established theories in social and forensic psychology are used to represent the news at each level. They utilize a supervised machine learning framework, deploying several supervised classifiers with five-fold cross-validation, namely, Support Vector Machine (with linear kernel), Random Forest (RF), and XGBoost10~\cite{xgboost}. They used well established datasets of news articles that were collected from PolitiFact and BuzzFeed~\cite{defend,shu2019fakenewsnet,shu2018news}.
\section{Conclusion and Future Works}
\label{sec-conclusion}

The presence of fake news in the internet is a rapidly growing problem that has engendered a lot of effort in fake news detection approaches within the machine learning community. While there have been many machine learning techniques in fake news detection, there is a need to analyze the robustness of these techniques in the presence of noise, which is very likely to exist in real-world news content. This work is an important step in this direction, where we evaluated the impact of noise on a fake news detection model, Hierarchical Discourse-level Structure for Fake News Detection Model or HDSF~\cite{karimi}, that has been found to outperform other state-of-the-art models. 

In future, evaluation studies for understanding the memorization and generalization properties of HDSF and other models in fake news detection can be carried out, as has been done in other domains, such as the study by Arpit et al.~\cite{arpit} which leveraged findings based on training their models with noise-induced image datasets. Such studies are likely to shed more light on the learning behavior of such models. Another direction is to build and train more sophisticated models with richer datasets that can predict what parts of a news article are potentially fake. This could be beneficial for interpreting how a model reached an authenticity conclusion about an article. Lastly, it is interesting to investigate the potential of HDSF in performing prediction tasks in other domains that involve structured data, such as software code.

\bibliographystyle{ACM-Reference-Format}
\bibliography{sample-base}


\begin{thebibliography}{23}


\ifx \showCODEN    \undefined \def \showCODEN     #1{\unskip}     \fi
\ifx \showDOI      \undefined \def \showDOI       #1{#1}\fi
\ifx \showISBNx    \undefined \def \showISBNx     #1{\unskip}     \fi
\ifx \showISBNxiii \undefined \def \showISBNxiii  #1{\unskip}     \fi
\ifx \showISSN     \undefined \def \showISSN      #1{\unskip}     \fi
\ifx \showLCCN     \undefined \def \showLCCN      #1{\unskip}     \fi
\ifx \shownote     \undefined \def \shownote      #1{#1}          \fi
\ifx \showarticletitle \undefined \def \showarticletitle #1{#1}   \fi
\ifx \showURL      \undefined \def \showURL       {\relax}        \fi
\providecommand\bibfield[2]{#2}
\providecommand\bibinfo[2]{#2}
\providecommand\natexlab[1]{#1}
\providecommand\showeprint[2][]{arXiv:#2}

\bibitem[Arpit et~al\mbox{.}(2017)]%
        {arpit}
\bibfield{author}{\bibinfo{person}{Devansh Arpit}, \bibinfo{person}{Stanisław
  Jastrzębski}, \bibinfo{person}{Nicolas Ballas}, \bibinfo{person}{David
  Krueger}, \bibinfo{person}{Emmanuel Bengio}, \bibinfo{person}{Maxinder~S.
  Kanwal}, \bibinfo{person}{Tegan Maharaj}, \bibinfo{person}{Asja Fischer},
  \bibinfo{person}{Aaron Courville}, \bibinfo{person}{Yoshua Bengio}, {and}
  \bibinfo{person}{Simon Lacoste-Julien}.} \bibinfo{year}{2017}\natexlab{}.
\newblock \showarticletitle{A Closer Look at Memorization in Deep Networks}.
\newblock \bibinfo{journal}{\emph{arXiv preprint arXiv:1706.05394v2}}
  (\bibinfo{year}{2017}).
\newblock


\bibitem[Aslam et~al\mbox{.}(2021)]%
        {fdetect}
\bibfield{author}{\bibinfo{person}{Nida Aslam}, \bibinfo{person}{Irfan~Ullah
  Khan}, \bibinfo{person}{Farah~Salem Alotaibi},
  \bibinfo{person}{Lama~Abdulaziz Aldaej}, {and} \bibinfo{person}{Asma
  Khaled}.} \bibinfo{year}{2021}\natexlab{}.
\newblock \showarticletitle{Fake Detect: A Deep Learning Ensemble Model for
  Fake News Detection}.
\newblock \bibinfo{journal}{\emph{Complexity Journal}}  \bibinfo{volume}{2021},
  Article \bibinfo{articleno}{5557784} (\bibinfo{year}{2021}),
  \bibinfo{numpages}{33}~pages.
\newblock


\bibitem[Bach and Huang(2019)]%
        {bach19_interspeech}
\bibfield{author}{\bibinfo{person}{Nguyen Bach} {and} \bibinfo{person}{Fei
  Huang}.} \bibinfo{year}{2019}\natexlab{}.
\newblock \showarticletitle{{Noisy BiLSTM-Based Models for Disfluency
  Detection}}. In \bibinfo{booktitle}{\emph{Proc. Interspeech 2019}}.
  \bibinfo{pages}{4230--4234}.
\newblock
\urldef\tempurl%
\url{https://doi.org/10.21437/Interspeech.2019-1336}
\showDOI{\tempurl}


\bibitem[Chen and Guestrin(2016)]%
        {xgboost}
\bibfield{author}{\bibinfo{person}{Tianqi Chen} {and} \bibinfo{person}{Carlos
  Guestrin}.} \bibinfo{year}{2016}\natexlab{}.
\newblock \showarticletitle{XGBoost: A Scalable Tree Boosting System}. In
  \bibinfo{booktitle}{\emph{Proceedings of the 22nd ACM SIGKDD International
  Conference on Knowledge Discovery and Data Mining}}
  \emph{(\bibinfo{series}{KDD '16})}. \bibinfo{pages}{785–794}.
\newblock


\bibitem[Graves and Schmidhuber(2005)]%
        {phoneme}
\bibfield{author}{\bibinfo{person}{A. Graves} {and} \bibinfo{person}{J.
  Schmidhuber}.} \bibinfo{year}{2005}\natexlab{}.
\newblock \showarticletitle{Framewise phoneme classification with bidirectional
  LSTM networks}. In \bibinfo{booktitle}{\emph{Proceedings. 2005 IEEE
  International Joint Conference on Neural Networks, 2005.}},
  Vol.~\bibinfo{volume}{4}. \bibinfo{pages}{2047--2052 vol. 4}.
\newblock


\bibitem[Hakak et~al\mbox{.}(2021)]%
        {hakak}
\bibfield{author}{\bibinfo{person}{Saqib Hakak}, \bibinfo{person}{Mamoun
  Alazab}, \bibinfo{person}{Suleman Khan}, \bibinfo{person}{Thippa~Reddy
  Gadekallu}, \bibinfo{person}{Praveen Kumar~Reddy Maddikunta}, {and}
  \bibinfo{person}{Wazir~Zada Khan}.} \bibinfo{year}{2021}\natexlab{}.
\newblock \showarticletitle{An ensemble machine learning approach through
  effective feature extraction to classify fake news}.
\newblock \bibinfo{journal}{\emph{Future Generation Computer Systems}}
  \bibinfo{volume}{117} (\bibinfo{year}{2021}), \bibinfo{pages}{47--58}.
\newblock
\showISSN{0167-739X}


\bibitem[Horne et~al\mbox{.}(2019)]%
        {horne}
\bibfield{author}{\bibinfo{person}{Benjamin~D. Horne}, \bibinfo{person}{Jeppe
  N\o{}rregaard}, {and} \bibinfo{person}{Sibel Adali}.}
  \bibinfo{year}{2019}\natexlab{}.
\newblock \showarticletitle{Robust Fake News Detection Over Time and Attack}.
\newblock \bibinfo{journal}{\emph{ACM Trans. Intell. Syst. Technol.}}
  \bibinfo{volume}{11}, \bibinfo{number}{1}, Article \bibinfo{articleno}{7}
  (\bibinfo{year}{2019}).
\newblock


\bibitem[Karimi(2020)]%
        {hdfs-repo}
\bibfield{author}{\bibinfo{person}{Hamid Karimi}.}
  \bibinfo{year}{2020}\natexlab{}.
\newblock \bibinfo{title}{HDSF Source Repository}.
\newblock
\newblock
\urldef\tempurl%
\url{https://github.com/hamidkarimi/HDSF}
\showURL{%
\tempurl}


\bibitem[Karimi and Tang(2019)]%
        {karimi}
\bibfield{author}{\bibinfo{person}{Hamid Karimi} {and} \bibinfo{person}{Jiliang
  Tang}.} \bibinfo{year}{2019}\natexlab{}.
\newblock \showarticletitle{Learning Hierarchical Discourse-level Structure for
  Fake News Detection}. In \bibinfo{booktitle}{\emph{Proceedings of the 2019
  Conference of the North {A}merican Chapter of the Association for
  Computational Linguistics: Human Language Technologies, Volume 1 (Long and
  Short Papers)}}. \bibinfo{pages}{3432--3442}.
\newblock


\bibitem[Liu and Lapata(2018)]%
        {liu-lapata-2018-learning}
\bibfield{author}{\bibinfo{person}{Yang Liu} {and} \bibinfo{person}{Mirella
  Lapata}.} \bibinfo{year}{2018}\natexlab{}.
\newblock \showarticletitle{Learning Structured Text Representations}.
\newblock \bibinfo{journal}{\emph{Transactions of the Association for
  Computational Linguistics}}  \bibinfo{volume}{6} (\bibinfo{year}{2018}),
  \bibinfo{pages}{63--75}.
\newblock


\bibitem[Liu and Wu(2020)]%
        {fned}
\bibfield{author}{\bibinfo{person}{Yang Liu} {and}
  \bibinfo{person}{Yi-Fang~Brook Wu}.} \bibinfo{year}{2020}\natexlab{}.
\newblock \showarticletitle{FNED: A Deep Network for Fake News Early Detection
  on Social Media}.
\newblock \bibinfo{journal}{\emph{ACM Trans. Inf. Syst.}} \bibinfo{volume}{38},
  \bibinfo{number}{3}, Article \bibinfo{articleno}{25} (\bibinfo{year}{2020}),
  \bibinfo{numpages}{33}~pages.
\newblock


\bibitem[Ma et~al\mbox{.}(2016)]%
        {microblogs}
\bibfield{author}{\bibinfo{person}{Jing Ma}, \bibinfo{person}{Wei Gao},
  \bibinfo{person}{Prasenjit Mitra}, \bibinfo{person}{Sejeong Kwon},
  \bibinfo{person}{Bernard~J. Jansen}, \bibinfo{person}{Kam-Fai Wong}, {and}
  \bibinfo{person}{Meeyoung Cha}.} \bibinfo{year}{2016}\natexlab{}.
\newblock \showarticletitle{Detecting Rumors from Microblogs with Recurrent
  Neural Networks}. In \bibinfo{booktitle}{\emph{Proceedings of the
  Twenty-Fifth International Joint Conference on Artificial Intelligence}}
  \emph{(\bibinfo{series}{IJCAI'16})}. \bibinfo{publisher}{AAAI Press},
  \bibinfo{pages}{3818–3824}.
\newblock
\showISBNx{9781577357704}


\bibitem[Ma et~al\mbox{.}(2018)]%
        {ma-etal-2018-rumor}
\bibfield{author}{\bibinfo{person}{Jing Ma}, \bibinfo{person}{Wei Gao}, {and}
  \bibinfo{person}{Kam-Fai Wong}.} \bibinfo{year}{2018}\natexlab{}.
\newblock \showarticletitle{Rumor Detection on {T}witter with Tree-structured
  Recursive Neural Networks}. In \bibinfo{booktitle}{\emph{Proceedings of the
  56th Annual Meeting of the Association for Computational Linguistics (Volume
  1: Long Papers)}}. \bibinfo{publisher}{Association for Computational
  Linguistics}, \bibinfo{address}{Melbourne, Australia},
  \bibinfo{pages}{1980--1989}.
\newblock


\bibitem[Rabin et~al\mbox{.}(2021)]%
        {rabin2021memorization}
\bibfield{author}{\bibinfo{person}{Md~Rafiqul~Islam Rabin},
  \bibinfo{person}{Aftab Hussain}, \bibinfo{person}{Vincent~J. Hellendoorn},
  {and} \bibinfo{person}{Mohammad~Amin Alipour}.}
  \bibinfo{year}{2021}\natexlab{}.
\newblock \bibinfo{title}{Memorization and Generalization in Neural Code
  Intelligence Models}.
\newblock
\newblock
\showeprint[arxiv]{2106.08704}


\bibitem[Rubin and Lukoianova(2015)]%
        {truth-deception}
\bibfield{author}{\bibinfo{person}{Victoria~L. Rubin} {and}
  \bibinfo{person}{Tatiana Lukoianova}.} \bibinfo{year}{2015}\natexlab{}.
\newblock \showarticletitle{Truth and Deception at the Rhetorical Structure
  Level}.
\newblock \bibinfo{journal}{\emph{Journal of the Association for Information
  Science and Technology}} \bibinfo{volume}{66}, \bibinfo{number}{5}
  (\bibinfo{year}{2015}), \bibinfo{pages}{905–917}.
\newblock


\bibitem[Schuster and Paliwal(1997)]%
        {birnn}
\bibfield{author}{\bibinfo{person}{M. Schuster} {and} \bibinfo{person}{K.K.
  Paliwal}.} \bibinfo{year}{1997}\natexlab{}.
\newblock \showarticletitle{Bidirectional recurrent neural networks}.
\newblock \bibinfo{journal}{\emph{IEEE Transactions on Signal Processing}}
  \bibinfo{volume}{45}, \bibinfo{number}{11} (\bibinfo{year}{1997}),
  \bibinfo{pages}{2673--2681}.
\newblock


\bibitem[Shu et~al\mbox{.}(2019a)]%
        {defend}
\bibfield{author}{\bibinfo{person}{Kai Shu}, \bibinfo{person}{Limeng Cui},
  \bibinfo{person}{Suhang Wang}, \bibinfo{person}{Dongwon Lee}, {and}
  \bibinfo{person}{Huan Liu}.} \bibinfo{year}{2019}\natexlab{a}.
\newblock \showarticletitle{DEFEND: Explainable Fake News Detection}. In
  \bibinfo{booktitle}{\emph{Proceedings of the 25th ACM SIGKDD International
  Conference on Knowledge Discovery and Data Mining}}
  \emph{(\bibinfo{series}{KDD '19})}. \bibinfo{pages}{395–405}.
\newblock


\bibitem[Shu et~al\mbox{.}(2019b)]%
        {shu2019fakenewsnet}
\bibfield{author}{\bibinfo{person}{Kai Shu}, \bibinfo{person}{Deepak
  Mahudeswaran}, \bibinfo{person}{Suhang Wang}, \bibinfo{person}{Dongwon Lee},
  {and} \bibinfo{person}{Huan Liu}.} \bibinfo{year}{2019}\natexlab{b}.
\newblock \showarticletitle{FakeNewsNet: A Data Repository with News Content,
  Social Context and Spatial Temporal Information for Studying Fake News on
  Social Media}.
\newblock \bibinfo{journal}{\emph{arXiv preprint arXiv:1809.01286}}
  (\bibinfo{year}{2019}).
\newblock


\bibitem[Shu et~al\mbox{.}(2018)]%
        {shu2018news}
\bibfield{author}{\bibinfo{person}{Kai Shu}, \bibinfo{person}{Suhang Wang},
  {and} \bibinfo{person}{Huan Liu}.} \bibinfo{year}{2018}\natexlab{}.
\newblock \showarticletitle{Beyond News Contents: The Role of Social Context
  for Fake News Detection}.
\newblock \bibinfo{journal}{\emph{arXiv preprint arXiv:1712.07709}}
  (\bibinfo{year}{2018}).
\newblock


\bibitem[Wang(2017)]%
        {wang2017liar}
\bibfield{author}{\bibinfo{person}{William~Yang Wang}.}
  \bibinfo{year}{2017}\natexlab{}.
\newblock \showarticletitle{"Liar, Liar Pants on Fire": A New Benchmark Dataset
  for Fake News Detection}.
\newblock \bibinfo{journal}{\emph{arXiv preprint arXiv:1705.00648}}
  (\bibinfo{year}{2017}).
\newblock


\bibitem[Wikipedia(2022)]%
        {wiki-bilstm}
\bibfield{author}{\bibinfo{person}{Wikipedia}.}
  \bibinfo{year}{2022}\natexlab{}.
\newblock \bibinfo{title}{https://paperswithcode.com/method/bilstm}.
\newblock
\newblock
\urldef\tempurl%
\url{https://en.wikipedia.org/wiki/Bidirectional_recurrent_neural_networks}
\showURL{%
\tempurl}
\newblock
\shownote{accessed on Jan 1 2022}.


\bibitem[Zhang et~al\mbox{.}(2021)]%
        {still-rethink-gen}
\bibfield{author}{\bibinfo{person}{Chiyuan Zhang}, \bibinfo{person}{Samy
  Bengio}, \bibinfo{person}{Moritz Hardt}, \bibinfo{person}{Benjamin Recht},
  {and} \bibinfo{person}{Oriol Vinyals}.} \bibinfo{year}{2021}\natexlab{}.
\newblock \showarticletitle{Understanding Deep Learning (Still) Requires
  Rethinking Generalization}.
\newblock \bibinfo{journal}{\emph{Commun. ACM}} \bibinfo{volume}{64},
  \bibinfo{number}{3} (\bibinfo{date}{feb} \bibinfo{year}{2021}),
  \bibinfo{pages}{107–115}.
\newblock


\bibitem[Zhou et~al\mbox{.}(2020)]%
        {fake-news-theory}
\bibfield{author}{\bibinfo{person}{Xinyi Zhou}, \bibinfo{person}{Atishay Jain},
  \bibinfo{person}{Vir~V. Phoha}, {and} \bibinfo{person}{Reza Zafarani}.}
  \bibinfo{year}{2020}\natexlab{}.
\newblock \showarticletitle{Fake News Early Detection: A Theory-Driven Model}.
\newblock \bibinfo{journal}{\emph{Digital Threats: Research and Practice}}
  \bibinfo{volume}{1}, \bibinfo{number}{2}, Article \bibinfo{articleno}{12}
  (\bibinfo{year}{2020}).
\newblock


\end{thebibliography}


\end{document}